\documentclass[10pt]{article}
\usepackage{amsmath,amsthm,amssymb,amsfonts,cancel}
\usepackage{thmtools,enumitem}
\usepackage{algorithm}
\usepackage{mathtools}
\usepackage{dsfont}
\usepackage{bbm}
\usepackage{tensor}
\usepackage{bm}
\usepackage{graphicx}
\usepackage{multirow}
\usepackage{subfigure}
\usepackage{comment}
\usepackage{afterpage}
\usepackage{float}

\newcommand{\ignore}[1]{}

\mathchardef\mhyphen="2D % Define a "math hyphen"
\DeclareMathOperator*{\argmin}{\arg\!\min}

\let\originalleft\left
\let\originalright\right
\renewcommand{\left}{\mathopen{}\mathclose\bgroup\originalleft}
\renewcommand{\right}{\aftergroup\egroup\originalright}

\usepackage[numbers]{natbib}
\usepackage{comment}

\usepackage{graphicx}
\usepackage{hyperref}
\usepackage{url}
\usepackage{booktabs}
\usepackage{bbm}
\usepackage{soul,multicol}
\usepackage{subfigure}
\newtheorem{theorem}{Theorem}[section]

\usepackage[margin=1in]{geometry}
\usepackage{adjustbox}
\usepackage{url}
\newtheorem{definition}{Definition}%

\begin{document}
	\title{Cluster Explanation via Polyhedral Descriptions}
	\author{Connor Lawless \quad  Oktay G\"unl\"uk \\ School of Operations Research and Information Engineering, Cornell University}
	\date{\today}

\maketitle


\abstract{Clustering is an unsupervised learning problem that aims to partition unlabelled data points into groups with similar features. Traditional clustering algorithms provide limited insight into the groups they find as their main focus is accuracy and not the interpretability of the group assignments. This has spurred a recent line of work on explainable machine learning for clustering. In this paper we focus on the cluster description problem where, given a dataset and its partition into clusters, the task is to explain the clusters. We introduce a new approach to explain clusters by constructing polyhedra around each cluster while minimizing either the complexity of the resulting polyhedra or the number of features used in the description. We formulate the cluster description problem as an integer program and present a column generation approach to search over an exponential number of candidate half-spaces that can be used to build the polyhedra. To deal with large datasets, we  introduce a novel grouping scheme that first forms smaller groups of data points and then builds the polyhedra  around the grouped data, a strategy which out-performs simply sub-sampling data. Compared to state of the art cluster description algorithms, our approach is able to achieve competitive interpretability with improved description accuracy.}

\section{Introduction}

Machine learning (ML) is becoming an omnipresent aspect of the digital world. While ML systems are increasingly automating tasks such as image tagging or recommendations, there is increasing demand to use them as decision support tools in a number of settings such as criminal justice \citep{berk2012criminal, rudin2018optimized, zavrvsnik2021algorithmic}, medicine \citep{rajkomar2019machine, ustun2016supersparse, varol2017hydra}, and marketing \citep{dzyabura2018machine, hair2021data, ma2020machine}. Thus it is becoming increasingly critical that human users leveraging these ML tools understand and critique the outputs of the ML models to trust and act upon the recommendations. This is especially true for clustering, an unsupervised machine learning task, where a set of unlabelled data points are partitioned into groups \citep{xu2005survey}. Clustering is often used in industry as a tool to find sub-populations in a dataset such as customer segments \citep{kansal2018customer}, different media genres \citep{daudpota2019video}, or even patient subgroups in clinical studies \citep{wang2020unsupervised}. In these settings, practitioners often care less about the actual cluster assignments (i.e. which user is in which group) but rather a description of the groups found (i.e. a segment of users that consistently buy certain kinds of products). Unfortunately, many clustering algorithms only output cluster assignments, forcing users to work backwards to construct cluster descriptions. 

This paper focuses on the cluster description problem, where a fixed clustering partition of a set of data points with real or integer coordinates is given and the goal is to find a compact description of the clusters. This problem occurs naturally in a number of settings where a clustering has already been performed either by a black-box system, or on unseen or complex data (for example a graph structure) and needs to be subsequently explained using features that may not have even been used in the initial cluster assignment. Existing work on cluster description has primarily focused on using interpretable supervised learning approaches to predict cluster labels \citep{jain1999data}. However, these approaches focus primarily on the accuracy of the explanation and do not explicitly optimize the interpretability of the explanation.

\begin{figure}[!tb]
\centering
\begin{minipage}{.6\textwidth}
  \centering
    \begin{subfigure}
    \centering
      \includegraphics[width=0.3\textwidth]{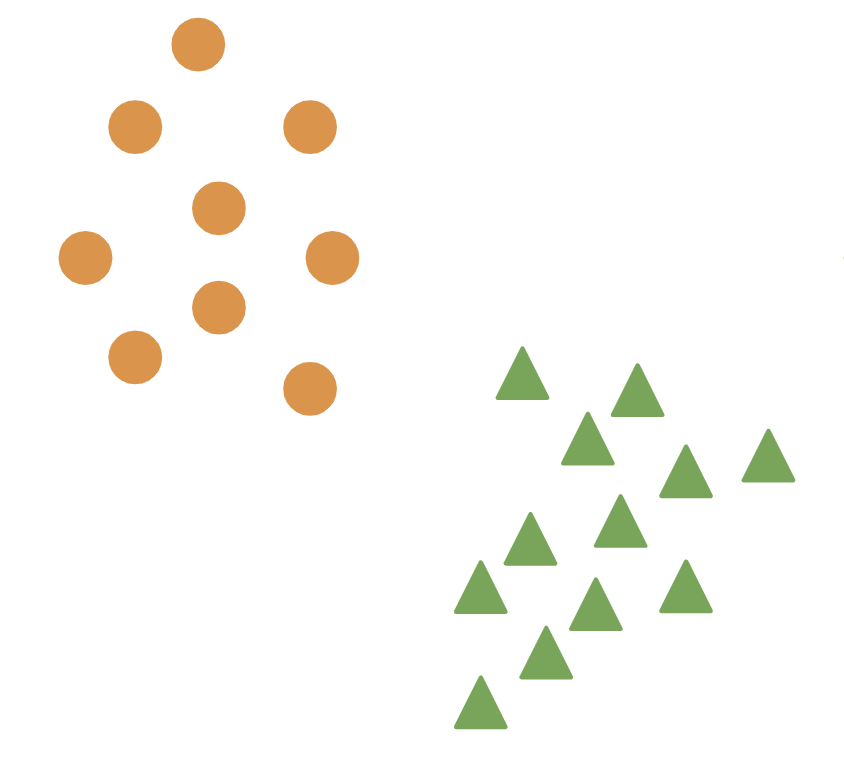}
    \end{subfigure}
    \begin{subfigure}
    \centering
      \includegraphics[width=0.3\textwidth]{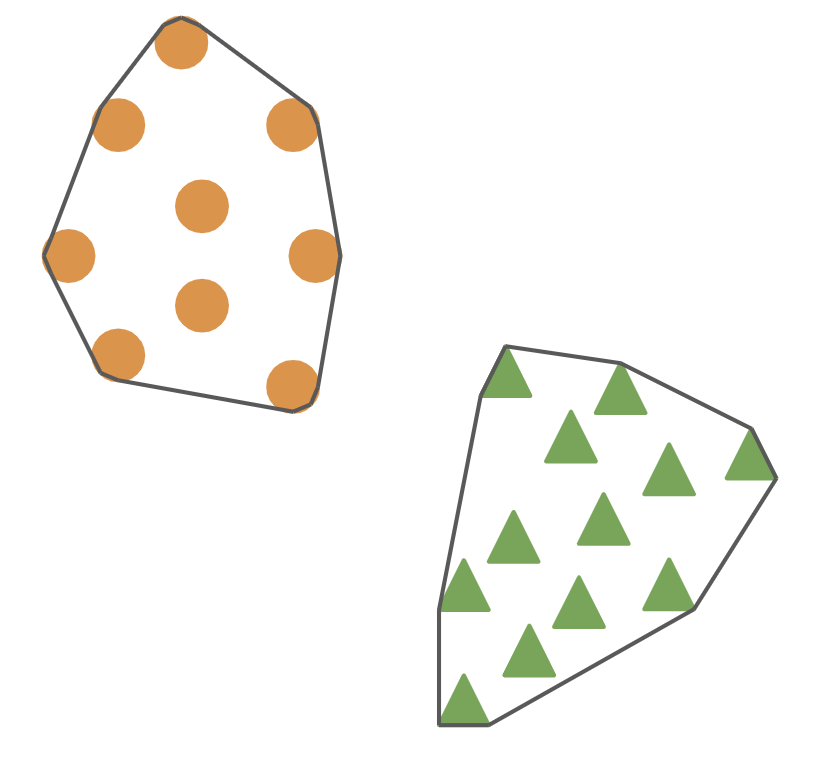}
    \end{subfigure}
    \begin{subfigure}
    \centering
      \includegraphics[width=0.3\textwidth]{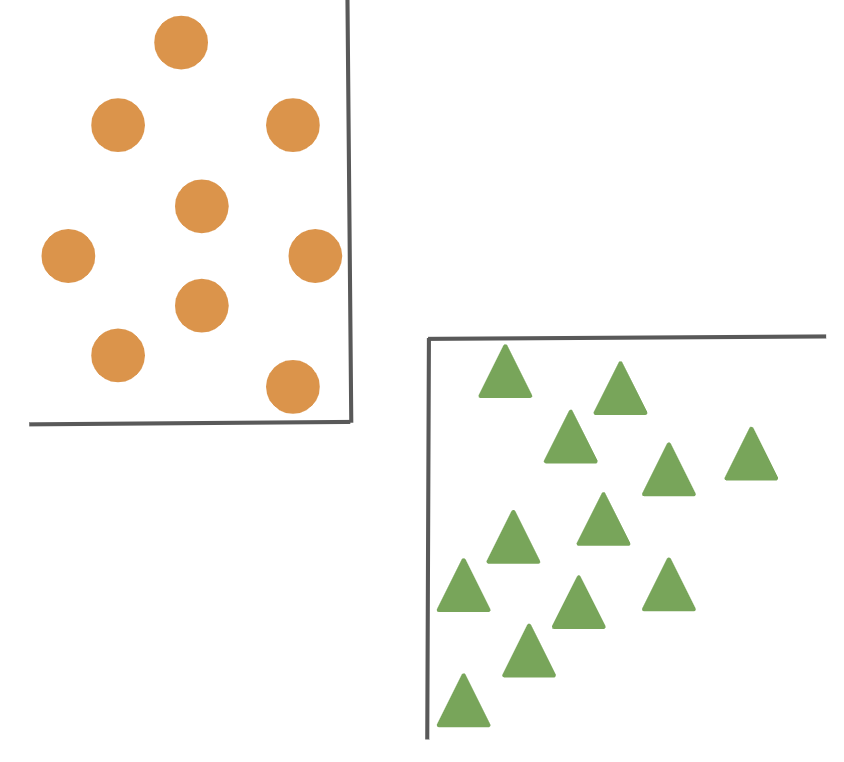}
    \end{subfigure}
  \caption{(Left) A sample set of clusters to be explained where the convex hulls do not intersect and perfect explanation is possible. (Middle) Polyhedral description using convex hull of clusters. (Right) Lower complexity polyhedral description.}
  \label{sample_polyhedra}
\end{minipage}~~~~~%~~~~
\begin{minipage}{.4\textwidth}
  \centering
    \begin{subfigure}
    \centering
      \includegraphics[width=0.48\textwidth]{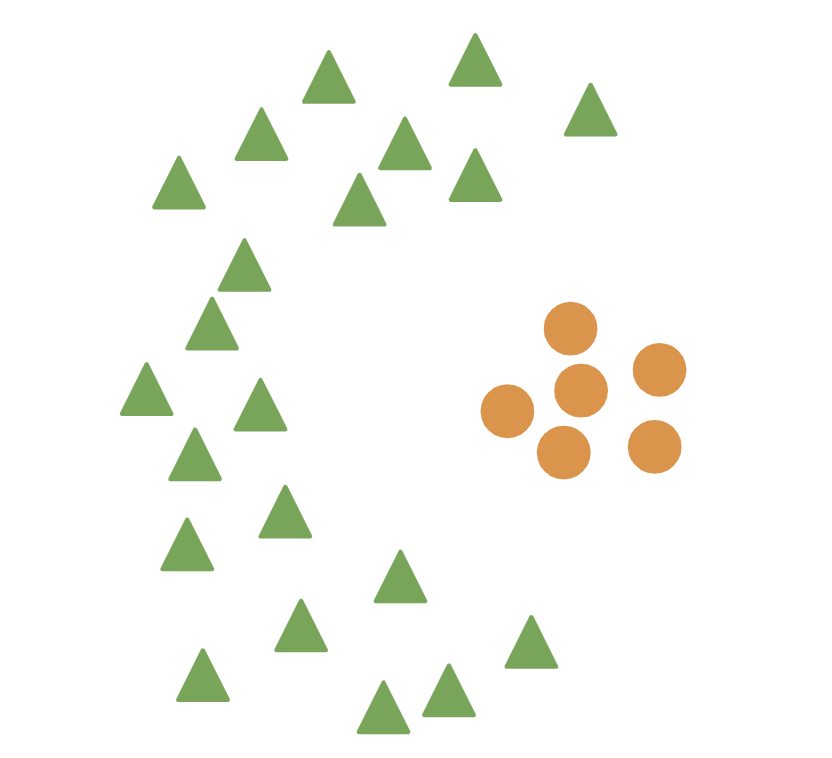}
    \end{subfigure}
    \begin{subfigure}
    \centering
      \includegraphics[width=0.40\textwidth]{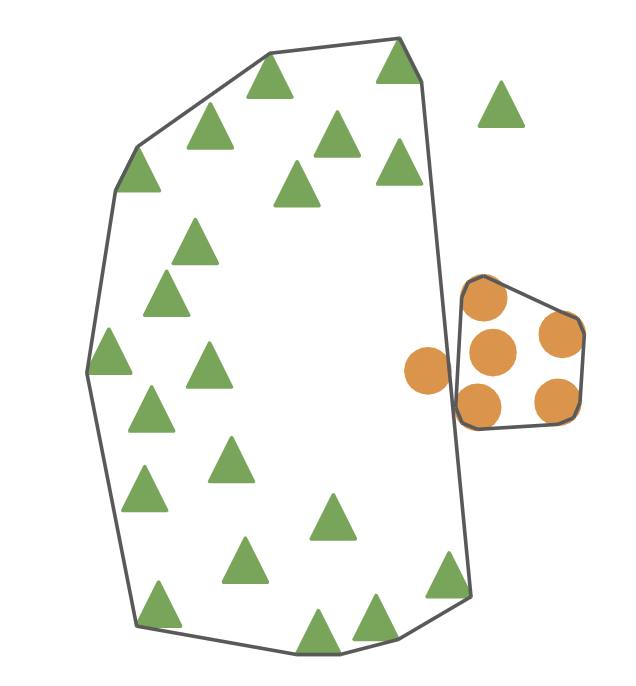}
    \end{subfigure}
  \caption{(Left) Sample set of clusters where convex hulls intersect and perfect explanation impossible. (Right) Best polyhedral description for clusters where convex hull intersects}
  \label{sample_bad_polyhedra}
\end{minipage}
\end{figure}

%constructing polyhedra around the given clusters to act as explanations
In this paper we introduce a new method for cluster description that treats each data point as a vector in $\mathbb{R}^n$ and works by constructing a polyhedron around each cluster to act as its explanation, henceforth referred to as \textit{polyhedral descriptions}. 
Each polyhedron is obtained by intersecting a (small) number of half spaces. We measure the interpretability of these polyhedra using two different notions: \emph{complexity}, which is defined to be the number of half-spaces used plus the sum of the number of nonzero coefficients used to define each half-space, or \emph{sparsity}, which is defined to be the number of features used across all half-spaces defining the polyhedra. If the convex hulls of the data points in each cluster do not intersect, then the half spaces defining the convex hull of the points in a cluster gives a polyhedral description for the cluster. However, such polyhedra may not have desirable interpretability characteristics as it might require a large number of half-spaces or involve many features. In this case a simpler explanation with some error might be more desirable. Furthermore, if the convex hulls of the clusters intersect then no error-free polyhedral description exists. Figures \ref{sample_polyhedra} and \ref{sample_bad_polyhedra} show examples of the polyhedra associated with both cases.

In our setting, the accuracy of a cluster explanation is measured by the fraction of data points that are correctly explained (i.e. included in the polyhedron of their cluster and not included in other polyhedra). This framework allows the cluster explanation to trade-off  accuracy with interpretability. Moreover, we can explicitly optimize the interpretability of the final polyhedral descriptions with respect to both complexity and sparsity. While a polyhedron may not initially seem like an interpretable model class, additional constraints placed on the half-spaces that construct the polyhedron allows cluster descriptions from popular interpretable model classes such as rule sets \citep{lawless2021interpretable, rudin2018learning, wang2017bayesian} and score cards \citep{ustun2017optimized}.

We formulate the problem of describing clusters with polyhedra as an integer program (IP) with exponentially many variables, one for each possible half-space. To solve this exponential size IP we present a column-generation algorithm that searches over candidate half-spaces without explicitly enumerating them. Using IP approaches in ML is known to be challenging due to the size of data sets which lead to computationally intractable IP instances. To deal with this scalability problem we introduce a novel grouping scheme where we first group small batches of data points together and perform the explanation on the grouped data allowing our approach to leverage information from the entire dataset while still scaling to larger datasets. Empirically, we demonstrate that this approaches out-performs simply sub-sampling the dataset.

\subsection{Related Work}

Existing work in interpretable clustering can be broadly divided into two groups: \emph{cluster description}, where cluster assignments are given and the task is to explain them (our work builds on this line of research); or \emph{interpretable clustering} approaches, where cluster assignments are generated using an interpretable model class. 

%Cluster description problem with tags
A common approach for cluster description is to simply use a supervised learning algorithm to predict the cluster label assignments that are already given \citep{de2017explaining, jain1999data, kauffmann2019clustering}. Broadly this can be seen as the application of multi-class classification \citep{aly2005survey} to cluster description. However, multi-class classification and cluster description differ in a few important characteristics. In multi-class classification the objective is to maximize classification accuracy, whereas the aim of cluster description is to explain the given clusters as simply as possible. In other words, cluster description aims to optimize interpretability with constraints on accuracy. Multi-class classification models are also expected to perform inference (i.e. make a prediction on new data). In cluster description there is no guarantee that the explanation is a partition of the feature space, and thus new data points can possibly fall outside all existing cluster descriptions. 

In a recent work Carrizosa et al. \cite{CARRIZOSA2022102543} introduce an IP framework for selecting a single prototype data point from each cluster and build a ball around it to act as a description for the cluster. While selecting a prototype point has an intuitive appeal, the resulting explanation can be misleading or uninformative if clusters are not compact or isotropic (i.e. have equal variance in all directions). Davidson et al. \cite{davidson2018cluster} introduce a version of the cluster description problem where each data point has an associated set of tags coming from a discrete set. The goal of their formulation is to find a disjoint set of tags for each cluster such that each data point in a cluster is covered by at least one tag assigned to that cluster, which they call the disjoint-tag descriptor minimization problem (DTDM). If we interpret each half-space in a polyhedral description as a tag, our approach bares a superficial resemblance to the DTDM problem but it also differs in a number of ways. First, a data point satisfies a description in the polyhedral description setting only if it satisfies all the conditions in the description, whereas in the DTDM a data point only needs to satisfy one of the tags used to describe the cluster. Unlike the DTDM, our framework  does not require data be provided with discrete tags and allows for real valued features. Finally, a data point is not considered correctly described in the polyhedral description problem if it meets a description for another cluster, a constraint not included in the DTDM. We note that this constraint ensures cluster descriptions that are informative (i.e. describe only a single cluster).

%We note that the absence of this constraint in the DTDM can lead to arbitrary cluster descriptions.
%Consider a simple example where there are $K$ clusters, and at least $K$ possible tags where the first $K$ tags are met by every data point. It is easy to see that an optimal solution to this example is to simply uniquely assign one of the first $K$ tags to each cluster. However, any permutation of the cluster descriptions to different clusters is also valid, highlighting that the discovered cluster descriptions are arbitrary. %Note that this is feasible for the DTDM as the tags are disjoint (each tag is only used once), and each data point in each cluster contains a tag selected for the cluster. It is also optimal as a valid lower bound to the DTDM is $K$ (i.e. no solution can outperform a single tag per cluster). 

%Interpretable Clustering v
There has also been extensive work on constructing clusters using interpretable model classes. The important distinction between this line of work and our setting is that this line of research assumes that the cluster assignment is not fixed. The majority of work has focused on the use of decision trees with uni-variate splits to perform the clustering \citep{bertsimas2021interpretable,  fraiman2013interpretable, frost2020exkmc, liu2000, pmlr-v119-moshkovitz20a}, or rule sets \citep{CARRIZOSA2022102543, chen2018interpretable, chen2016interpretable, pelleg2001mixtures}. Dasgupta et al. \cite{dasgupta2020explainable} use k-means as a reference clustering algorithm then introduce a decision tree algorithm that tries to minimize any increases in the $k$-means clustering cost. The algorithm, Iterative Mistake Minimization (IMM), assigns one leaf node per cluster which gives similar descriptions to a polyhedron constructed from uni-variate hyperplanes. While this may seem similar to polyhedral cluster descriptions it is important to note that the objective is the k-means clustering cost, not explicitly interpretability, and operates only with a $k$-means reference clustering not a general clustering. Most similar to our work is the use of multi-polytope machines to perform the clustering \citep{lawless2021interpretable}. However, our approach differs from this line of work as the cluster assignments are fixed in the cluster description problem, and the aim is optimize interpretability not the quality of the clustering itself. The cluster description setting can also be modeled as an IP as opposed to a mixed-integer non-linear program (MINLP) which allows our approach to scale to larger datasets.

\subsection{Main Contributions}
We summarize our main contributions as follows:
\begin{itemize}
    \item We introduce the polyhedral description problem which aims to explain why the data points in the same cluster are grouped together by building a polyhedron around them. We also show that this is an NP-Hard problem.
    \item We formulate the polyhedral description problem as an (exponential size) integer program where variables correspond to candidate half-spaces that can be used in the polyhedral description of the clusters. We present a column-generation framework to search over the (exponentially many) candidate half-spaces efficiently.
    \item We introduce a novel grouping scheme to summarize input data. This approach helps reduce the size of the IP instances and enables us to handle  large datasets. We also present empirical results that show that our grouping scheme out-performs the commonly used approach of sub-sampling data points.
    \item We present numerical experiments on a number of real world clustering datasets and show that our approach performs favorably compared to state of the art cluster description approaches. 
\end{itemize}

The remainder of the paper is organized as follows. Section \ref{sec:prob_form} formalizes the polyhedral description problem and presents an exponential sized IP formulation for constructing optimal polyhedral descriptions together with a column generation approach for solving it. Section \ref{sec:group} introduces a novel grouping scheme to enable the IP approach to deal with large scale data. Finally, Section \ref{sec:exp} presents numerical results on a suite of UCI clustering data sets.

\section{Problem Formulation} \label{sec:prob_form}

We now formally introduce the Polyhedral Description Problem (PDP). The input data for the problem consists of a set of $n$ data points with $m$ real-valued features ${\cal X} = \{x^i \in \mathbb{R}^m \}_{i=1}^{N}$ and a partition of the data points in ${\cal X}$ into $K$ clusters $C_1, \dots, C_K$, where each $C_{k}$ denotes the set of data points belonging to cluster $k$. Let $x^i_d$ be the $d$-th feature of the data point $x^i$ and $k_i$ be its cluster assignment. For a given $w\in\mathbbm{R}^m$ and $b\in\mathbbm{R}$, the half-space associated with $(w,b)$ is the set $h = \{x \in \mathbbm{R}^m: w^Tx \leq b\}$. For the remainder of the paper we refer to the half-space and the hyperplane defining a half-space interchangeably (i.e. refer to $\|w\|_0$ as the number of features used in a half-space). A polyhedron is the intersection of a finite number of half-spaces. 

When it is possible, a solution to the PDP is a set of polyhedra $\{ P_k\}_{k=1}^K$ such that $x \in C_k$ for all $ x \in P_k$ and $x \notin C_k$ for all $ x \notin P_k$. 
When no such polyhedral description exists (i.e. the convex hulls of the clusters intersect), the goal is to find a solution to the PDP subject to a budget $\alpha$ on the number of data points incorrectly explained: $$\bigg| \big\{x^i \in {\cal X}: x^i \notin P_{k_i} \lor x^i \in \cup_{k'\neq k_i }P_{k'}\big\}\bigg| \leq \alpha$$ 

We say that a data point $x \in \cal X$ is correctly explained if $x \in P_{k_i}$ and $x \notin \cup_{k'\neq k_i }P_{k'}$. To improve the interpretability of the resulting descriptions we consider a restricted set of candidate half-spaces ${\cal H}$ that are defined by sparse hyperplanes with small integer coefficients. More precisely, we consider half-spaces that have the form $\{x \in \mathbbm{R}^m : w^T x \leq b \}$ for integral $w$ with maximum value $W$, $\max_d |w_d| \leq W$, and at most $\beta$ non-zero values, $\|w\|_0 \leq \beta$. Note that additional restrictions on the set of candidate half-spaces may cause the PDP to be infeasible, even if the convex hulls of the points in each cluster do not intersect.

It is important to note that this approach does not require the polyhedra to be non-intersecting, but rather penalizes data points that fall into multiple polyhedra. From a practical perspective, adding such a restriction on the polyhedra would lead to a computationally challenging problem. It may also be overly restrictive in settings where the intersection of polyhedra is unlikely to contain any data (see the appendix for an illustrative example). In our computational experiments we observed only a small number of data points in the intersection of multiple polyhedra while there were many examples of polyhedra intersecting.

We consider two different variations of the PDP that add additional restrictions on the polyhedral descriptions to help improve interpretability. The first is to put a constraint on the complexity of the polyhedral description. Similar to previous work on rule sets \citep{lawless2021interpretable}, we define complexity of a half-space %\oo{$\{x: w^Tx \leq b\}$ as $\|w\|_0+1$. }
as the number of non-zero terms in the half-space plus one, and the complexity of the polyhedron as the sum of the complexities of the half-spaces that compose it. 
We call this  variant of the PDP with complexity constraint, the \emph{Low-Complexity} PDP (LC-PDP).
The second variant we consider puts a limit on the total number of features in all the half-spaces used in the polyhedral descriptions (i.e. sparsity). 
We call the second variant of the PDP with the sparsity constraint, the \emph{Sparse} PDP (Sp-PDP). The decision version of the PDP involves deciding if there exists a feasible polyhedral description subject to a constraint on complexity or sparsity whereas the optimization version of the PDP involves minimizing the complexity or sparsity of the solution. Unfortunately, the decision version of both variants of the Polyhedral Description Problem are NP-Hard. All proofs can be found in the Appendix.

\begin{theorem} \label{thm:np}
Both the Low Complexity and Sparse Polyhedral Description Problems are NP-Hard.
\end{theorem}

\subsection{Integer Programming Formulation for the PDP} \label{sec:IP}

Given a set of candidate half-spaces ${\cal H}$ that can be used in a polyhedral description, we next formulate the optimization version of both variants of the PDP as an integer program. In practice, enumerating all possible candidate half-spaces, even in this restricted setting, is computationally impractical and we discuss a column generation approach to handle this in the subsequent section. Let 
$${\cal H}_i = \{ (w,b) \in {\cal H}: w^T x^i > b\}$$ 
be the set of half-spaces that data point $i$ falls outside, and in a slight abuse of notation let 
$${\cal H}_d = \{ (w,b) \in {\cal H}: w_d \neq 0 \}$$ 
be the set of half-spaces that use feature $d$. For each half-space we define its complexity as the number of features used in the half-space plus a penalty of one. Formally the complexity $c_h$ for half-space $h = (w,b)$ is defined as $c_h = \|w\|_0 + 1$.

Let $z_{hk}$ be the binary decision variable indicating whether half-space $h$ is used in the polyhedral description for cluster $k$. Note that we can recover the polyhedral description for cluster $k$ from these binary variables as 
$$P_k = \bigcap_{h \in I_k} h~~\text{where}~~I_k = \{h \in {\cal H}: z_{hk} = 1\}.$$ 
We use a binary variable $\xi_i$ to indicate whether data point $i$ is mis-classified (i.e. either not included in the cluster's polyhedron or is incorrectly included in another cluster's polyhedron). Let $y_d$ be a binary variable indicating whether feature $d$ is used in any of the half-spaces chosen for the polyhedral descriptions. With these definitions, an integer programming formulation for the PDP is as follows:

\begin{align}
	\hskip3cm\textbf{min}~~&& \theta_1 \sum_{k=1}^K \sum_{h \in {\cal H}} c_h z_{hk} &+ \theta_2 \sum_{d = 1}^m y_d \label{obj:mip}\\
	\textbf{s.t.}~~&& \xi_i + \sum_{h \in {\cal H}_{i}} z_{hk} &\geq 1 ~~&&\forall x^i \in {\cal X}, \forall k \neq k_i \hskip4cm\label{const:FalsePos}\\
	&& M\xi_i - \sum_{h \in {\cal H}_{i}} z_{hk_i} &\geq 0 ~~ &&\forall x^i \in {\cal X} \label{const:FalseNeg}\\
	&& \sum_{k=1}^K \sum_{h \in {\cal H}_d} z_{hk} &\leq M y_d &&\forall d \in \{1,\dots,m\} \label{const:sparsityD}\\ 
	&&\sum_{x^i \in {\cal X}} \xi_i &\leq \alpha \label{const:error_bound} \\
	%        \sum_{d} y_d \leq \beta \label{const:sparsity}\\
	&&\xi_i, z_{hk}, y_d &\in \{0,1\} \label{const:binary}
\end{align}

 %  \begin{align}
%	\min \theta_1 \sum_{k=1}^K \sum_{h \in {\cal H}} c_h z_{hk} &+ \theta_2 \sum_{d = 1}^m y_d  \label{obj:mip}\\
%	\textbf{s.t.}~~ \xi_i + \sum_{h \in {\cal H}_{i}} z_{hk} &\geq 1 ~~ \forall x^i \in {\cal X}, \forall k \neq k_i \label{const:FalsePos}\\
%	M\xi_i - \sum_{h \in {\cal H}_{i}} z_{hk_i} &\geq 0 ~~ \forall x^i \in {\cal X} \label{const:FalseNeg}\\
%	\sum_{k=1}^K \sum_{h \in {\cal H}_d} z_{hk} &\leq M y_d ~\forall d \in \{1,\dots,m\} \label{const:sparsityD}\\ 
%	\sum_{x^i \in {\cal X}} \xi_i &\leq \alpha \label{const:error_bound} \\
%	%        \sum_{d} y_d \leq \beta \label{const:sparsity}\\
%	\xi_i, z_{hk}, y_d &\in \{0,1\} \label{const:binary}
%\end{align}
%
     
    The objective consists of two terms that capture both variants of the PDP. The first term captures the complexity of the half-spaces used (LC-PDP), and the second captures the sparsity (Sp-PDP). $\theta_1$ and $\theta_2$ control the relative importance of each term. Note that if $\theta_1=1, \theta_2 = 0$ we get the LC-PDP, and similarly if $\theta_1=0, \theta_2 = 1$ we get the Sp-PDP. 
    
    %A natural choice is the smallest upper bound for the total number of half-spaces used (if an existing heuristic solution exists), or simply $|{\cal H}_i|$.
    
    Constraint \eqref{const:FalsePos} tracks false positives (i.e. data points that are included in a wrong cluster's polyhedron) and constraint \eqref{const:FalseNeg} tracks false negatives (i.e. data points that are not included in their respective cluster's polyhedron). Constraint \eqref{const:sparsityD} tracks which features are used in the polyhedral descriptions. If $\theta_2 = 0$ (i.e. sparsity is not a consideration) then constraint \eqref{const:sparsityD} can be removed and the problem can be decomposed into a separate problem for each cluster. In constraint \eqref{const:sparsityD} $M$ is a suitably large constant such as an upper bound on the objective value. Note that in practice the choice of $M$ can be chosen independently for constraints \eqref{const:FalseNeg} and \eqref{const:sparsityD}. Constraint \eqref{const:error_bound} sets an upper bound $\alpha$ on the number of data points that are not properly explained. We denote the problem \eqref{obj:mip}-\eqref{const:binary} as the master integer program (MIP), and its associated linear relaxation, taken by relaxing constraint \eqref{const:binary} to allow for non-integer values, as the master LP (MLP).

\subsection{Column Generation}

Enumerating every possible half-space is computationally intractable and thus it is not practical to solve the MIP using standard branch-and-bound techniques \citep{LandDoig}. Instead, we use column generation \citep{GilmoreGomory} to solve the MLP by searching over the best possible candidate half-spaces to consider in the master problem. Once we solve the MLP to (near) optimality or exceed a computational budget, we then use the set of candidate half-spaces generated during column generation to find a solution to the MIP. To solve the MLP we start with a restricted initial set of half-spaces $\hat{{\cal H}} \subset {\cal H}$. We denote the MLP solved using only $\hat{{\cal H}}$ the restricted master linear program (RMLP). In other words, the RMLP is the MLP where all variables corresponding to ${\cal H} \setminus \hat{{\cal H}}$ are set to 0. Once this small instance of the MLP is solved, we use the optimal \emph{dual} solutions to the problem to identify a missing variable (i.e. half-space) that has a negative reduced cost. The problem to find such a half-space is called the \emph{pricing problem} and can be solved by another integer program. If a new half-space with a negative reduced cost is found than we add it to the set $\hat{{\cal H}}$ and this process is repeated again until either no such half-space can be found, which represents a certificate of optimality for the MLP, or a given computational budget is exceeded. 

Let $(\mu, \gamma, \phi)$ be the optimal dual solution to the RMLP where $\mu_{ik} \geq 0$ is the dual value corresponding to constraint \eqref{const:FalsePos} for data point $i$ and cluster $k$, $\gamma_i \geq 0$ is the dual value corresponding to constraint \eqref{const:FalseNeg} for data point $i$, and $\phi_d \leq 0$ is the dual value corresponding to constraint \eqref{const:sparsityD} for dimension $d$, respectively. Since the decision variables $z_{hk}$ in the MIP are defined for a half-space and a specific cluster $k$, we define a separate pricing problem for each cluster, which can be solved in parallel. Using the optimal dual solution, the reduced cost $\rho_{(h,k)}$ for a missing variable $z_{hk}$ corresponding to a half-space $h \notin \hat{{\cal H}}$ for a cluster $k$ is:
$$
\rho_{(h,k)} = \theta_1 c_h - \sum_{i \in {\cal X} \setminus C_k} \mu_{ik} \mathbbm{1}(w^T x_i > b) +  \sum_{i \in C_k} \gamma_i \mathbbm{1}(w^T x_i > b) - \sum_{d=1}^m \phi_d \mathbbm{1}(\text{$w_d \neq 0$})
$$
Where $\mathbbm{1}(x)$ is the indicator function and equals $1$ if the literal $x$ is true, and 0 otherwise. Note that $\rho_{(h,k)} \geq 0 ~~\forall h \in \hat{{\cal H}}$ by the optimality of the dual solution. For a given cluster $k$ let $w \in \mathbbm{Z}^m$ and $b \in \mathbbm{R}$ be the decision variables representing the hyperplane used to construct a candidate half-space. We also introduce variables $w^+, w^- \in \mathbbm{Z}_{\geq 0}$ that represent the positive and negative components of the hyperplane (i.e. $w^+_d = \max(0, w_d)$ and $w^-_d = \max(0, -w_d)$). Let $y_d$ be the binary variable indicating whether feature $d$ is used in the hyperplane, and similarly $y_d^+, y_d^-$ represent whether a positive or negative component of feature $d$ is used. Finally let $\delta_i$ be the binary variable indicating whether data point $x^i \in {\cal X}$ is correctly included, for data points in $C_k$, or excluded, for data points in ${\cal X} \setminus C_k$, in the half-space. With these decision variables in mind, the pricing problem to find a candidate half-space for cluster $k$ can be formulated as follows:

   \begin{align}
         \hskip1.3cm\textbf{min}~~&&\theta_1(\sum_{d=1}^m (y_d^+ + y_d^-) + 1)
         - \sum_{x^i \in {\cal X} \setminus C_k} \mu_{ik} (1 - \delta_i) + \sum_{x^i \in C_k} \gamma_i \delta_i - \sum_{d=1}^m \phi_d (y_d^+ + y_d^-)\hskip.6cm\\[-.85cm]\nonumber
\end{align}
     \begin{align}
  %       \hskip1.8cm\textbf{min}~~&&\theta_1(\sum_{d=1}^m (y_d^+ + y_d^-) + 1)-& \sum_{x^i \in {\cal X} \setminus C_k} \mu_{ik} (1 - \delta_i) + \sum_{x^i \in C_k} \gamma_i \delta_i &&- \sum_{d=1}^m \phi_d (y_d^+ + y_d^-)\\
        \hskip1cm\textbf{s.t.}~~&& (w^+ - w^-)^T x^i - b &\leq M\delta_i ~~ &&\forall x^i \in C_k  \label{const:halfspace_error_1}\\
        &&(w^+ - w^-)^T x^i - b &\geq \epsilon -M \delta_i  ~~ ~~&&\forall x^i \in {\cal X} \setminus C_k  \label{const:halfspace_error_2}\\
        &&y_d^+ \leq w_d^+ &\leq W y_d^+ ~~&&\forall d \in \{1,\dots,m\}\label{const:max_coeff_1} \\ 
        &&y_d^- \leq w_d^- &\leq W y_d^- ~~&&\forall d \in \{1,\dots,m\}\label{const:max_coeff_2}\\ 
        &&\sum_{d=1}^m (y_d^+ + y_d^-) &\leq \beta \label{const:sparse_hp}\\
        &&y_d^+ + y_d^- &\leq 1 ~~&&\forall d \in \{1,\dots,m\}\label{const:trivial_1}\\ 
        &&\sum_{d=1}^m (w_d^+ + w_d^-) &\geq 1 \label{const:trivial_2}\\ 
        &&w_d^+,w_d^- &\in \mathbbm{Z}_{\geq 0} ~~&&\forall d \in \{1,\dots,m\} \\
        &&y_d, \delta_i &\in \{0,1\} ~~&&\forall d \in \{1,\dots,m\},~ x^i \in {\cal X}\hskip1.5cm
    \end{align}

The objective of the problem is to minimize the reduced cost of the new column. Note that $c_h$ is defined by $\|w\|_0 + 1$ which can be represented by the $y_d$ variables in the objectives. Constraint \eqref{const:halfspace_error_1} tracks whether a data point in $C_k$ is included in the half-space and similarly Constraint \eqref{const:halfspace_error_2} tracks whether or not each data point outside of $C_k$ is not included in the half-space. M is a suitably large constant that can be computed based on the data set and settings for $W, \beta$. In the latter constraint $\epsilon$ is a small constant to ensure the constraint is a strict inequality. Constraints \eqref{const:max_coeff_1} and \eqref{const:max_coeff_2} put a bound on the maximum integer coefficient size of the hyperplane, and constraint \eqref{const:sparse_hp} puts a bound on the $\ell_0$ norm of the hyperplane. Finally, constraints \eqref{const:trivial_1} and \eqref{const:trivial_2} exist to exclude the trivial solution where $w = 0$.

\section{Grouped Data for Scalability} \label{sec:group}
For problems with a large number of data points it can be computationally challenging to solve the IP formulation introduced in the preceding section. A standard approach for clustering or cluster description for large datasets is to simply sub-sample data points to consider in the optimization problem (see \citep{CARRIZOSA2022102543} for an example of the approach). While this approach has intuitive appeal, it fails to leverage all the information present in the given problem. Instead, we use a novel technique where we create smaller groups of data points that we treat as a single entity and perform the cluster description on the grouped data. This approach also effectively reduces the size of the problem instance without fully discard any of the data. 

\subsection{Description Error in Grouped Data} \label{sec:group_approx}

Grouping data points can have ambiguous affects on the interpretability of the final solution (i.e. can lead to solutions that are simpler or more complex). Moreover,  it may come at a cost to the accuracy of the cluster description (i.e. how many data points are correctly explained). In this section we formalize the notion of grouping data points and present results on its impact on the accuracy of the resulting cluster description.

We start by partitioning each cluster $C_k$ into a set of smaller groups ${\cal G}_k$ where each data point is assigned to a single group, and define ${\cal G} = \cup_{k=1}^K {\cal G}_k$. The scheme by which the groups are constructed can be viewed as a separate clustering task that can be performed by a user's clustering algorithm of choice. In practice we found that using a hierarchical clustering algorithm with a bound on the maximal linkage of each group performed the best empirically. We say that a group $G \in {\cal G}$ is correctly explained if all data points $x \in G$ are correctly explained. Let $\textbf{P} = \{P_k\}_{k=1}^K$ be a solution to the PDP (i.e. a set of polyhedral descriptions). We define the \emph{true} cost 
$$COST(\textbf{P}) = \sum_{k=1}^K \sum_{x \in C_k} \mathbbm{1}((x \notin P_k) \lor (x \in \bigcup_{k' \neq k}P_{k'}))$$ 
to be the number of data points incorrectly explained by the solution. For simplicity we exclude the explicit dependence of the dataset ${\cal X}$ and the cluster assignments ${\cal C}$ from the inputs to the cost function, but both are evidently necessary in determining the number of incorrectly explained data points.

We define the grouped cost 
$$COST_{G}(\textbf{P}) = \sum_{k=1}^K \sum_{G \in {\cal G}_k}|G| \mathbbm{1}(\exists x \in G \text{ s.t. } (x \notin P_k) \lor (x \in \bigcup_{k' \neq k}P_{k'}))),$$ 
as the mis-classification cost of each group weighted by the size of the group. A natural corollary of this definition is that for any solution $\textbf{P}$ the grouped cost \emph{over-estimates} the true cost (i.e. $COST_{G}(\textbf{P}) \geq COST(\textbf{P})$). Let $\textbf{P}^*_{G} = \argmin_{\textbf{P}} COST_{G}(\textbf{P})$ and $\textbf{P}^* = \argmin_{\textbf{P}} COST(\textbf{P})$ be the optimal solutions to the grouped problem and original problem respectively. We now show that solving the PDP over groups versus the individual data points leads to mis-classifying at most $|G_{max}|$ times the optimal number of data points, where $|G_{max}|$ is the size of the largest group.

\begin{theorem} \label{thm:group_ub}
The optimal solution to the grouped problem, with \textbf{any grouping scheme}, incurs a cost no more than $|G_{max}|$ times the cost of the optimal solution to the full problem instance. Formally:
$$
COST(\textbf{P}^*_G) \leq |G_{max}|COST(\textbf{P}^*) 
$$
\end{theorem}
While $|G_{max}|$ may seem like a relatively large cost, it is important to note that even creating small groups can have large impacts on the size of problem instances that can be solved via integer programming (i.e. even groups of size 2 halves the IP instance size). One important distinction about Theorem \ref{thm:group_ub} is that it places no assumption on how the groups were formed (i.e. the grouping scheme), and thus provides a general bound for any grouping approach. A natural question is whether placing additional restrictions on how groups are formed can lead to a stronger guarantee. One such possible restriction is to ensure that the grouping is optimal with respect to a clustering evaluation metric. Silhouette coefficient is a popular clustering evaluation metric that has been used in a line of recent work on optimal interpretable clustering \citep{bertsimas2021interpretable, lawless2021interpretable}.

\begin{definition}[Silhouette Coefficient] Consider data point $x^i \in C_k$, and a distance matrix $d$ where entries $d_{ij}$ capture distance between data point $x^i$ and $x^j$. Let $r(x^i)$ be the average distance between data point $x^i$ and every other data point in the same cluster. Let $q(x^i)$ be the average distance between data point $x^i$ and every data point in the second closest cluster. For data point $x^i$ the silhouette score  $s(x^i)$ is defined as:
\begin{align*}
r(x^i) = \frac{1}{|C_k|-1}\sum_{x^j \in C_k}d_{ij} ~~~
q(x^i) = \min_{l = 1, \dots, K: l \neq k} \frac{1}{|C_l|}\sum_{x^j \in C_l}d_{ij} ~~~
s(x^i) = \frac{q(x^i) - r(x^i)}{\max(q(x^i),r(x^i))}
\end{align*}

The silhouette score for a set of cluster assignments is the average of the silhouette scores for all the data points. The possible values range from -1 (worst) to +1 (best).
\end{definition}

Unfortunately, the following result shows that the bound in Theorem \ref{thm:group_ub} is tight in the sense that there exists an instance where the grouped cost is equal to $|G_{max}|$ times the optimal cost on the full problem even when a large number of groups are used via an optimal grouping scheme with respect to the silhouette coefficient.

\begin{theorem} \label{thm:group_lb}
Even for $|{\cal G}_k| = |C_k| - 2$ and an optimal grouping scheme with respect to silhouette coefficient, there exists an instance where:
$$
COST(\textbf{P}^*_G) = |G_{max}|COST(\textbf{P}^*) 
$$
\end{theorem}

Note that although this theorem uses silhoeutte coefficient, we believe that the same bound exists for any other cluster evaluation metric. The emphasis of this result is that even when groups are constructed in a reasonable manner, there still exists an instance where the upper bound is tight.

\subsection{Integer Programming Formulation with Grouping} \label{sec:group_ip}

We next describe is how to integrate the grouped data into the original IP formulation presented in Section \ref{sec:IP}. The goal of the approach is to summarize the information about each group in such a way that the resulting integer program scales linearly with the number of groups. For this purpose we start with constructing the smallest hyper-rectangle that contains all the data points in each group. Let $x_{G,d}^{H} = max_{x \in G} x_d$ and $x_{G,d}^{L} = min_{x \in G} x_d$ be the maximum and minimum value for coordinate $d$ for the points in group $G$. The hyper-rectangle $R_G$ for the group $G$ is defined as the set $$R_G = \big\{x \in \mathbbm{R}^m: x_{G,d}^H \geq x_d \geq x_{G,d}^L ~\forall d=1,\dots,m \big\}.$$ In our new formulation we consider a group to be mis-classified if any part of the hyper-rectangle is mis-classified. Note that this is a stronger condition than the previous section where a group is mis-classified if any data point is mis-classified. Consider a simple example where the group $G = \{(0,1), (1,0)\}$ and the half-space $h = \{x \in \mathbbm{R}^2: x_1 + x_2 \leq 1.5\}$. Both data points are included in the half-space but the associated rectangle is not. However, modelling the pricing problem to track whether each individual data point is correctly classified would not reduce the problem size of the pricing problem, eliminating the computational benefit of leveraging grouping. It is also worth noting this difference only occurs for non-axis parallel half-spaces (i.e. $\beta > 1$). 

Let $w_+$ and $w_-$ again represent the positive and negative components of the hyperplane (i.e. $w_{+,d} = \max(w_d, 0)$, $w_{-,d} = \max(-w_d, 0)$). A hyper-rectangle for group $G$ is fully inside a half-space $h = (w,b)$ (i.e. $R_G \subset h$) if the following condition holds:
$$
w_+^T (x_G^{H}) - w_-^T (x_G^{L}) \leq b 
$$
Note this is akin to ensuring the worst-case corner of the hyper-rectangle is within a given half-space. Similarly, a hyper-rectangle for a group $G$ is fully outside a half-space (i.e. $R_G \cap h = \emptyset $) if:

$$
w_+^T (x_G^{L}) - w_-^T (x_G^{H}) > b 
$$

We can now integrate the hyper-rectangle approach into the IP formulation as follows. In the master problem, let ${\cal H}_G^+ $ and ${\cal H}_G^-$ represent the set of half-spaces that group $G$ does not fully fall within or fall outside respectively. Formally
$${\cal H}_G^+ = \{ h \in {\cal H}: w_+^T (x_G^{H}) - w_-^T (x_G^{L}) > b \}$$ 
and 
$${\cal H}_G^- = \{ h \in {\cal H}: w_+^T (x_G^{L}) - w_-^T (x_G^{H}) > b  \}.$$ 
Constraints \eqref{const:FalsePos}, \eqref{const:FalseNeg}, and \eqref{const:error_bound} in the MLP/MIP are thus updated to the following:

\begin{align}
\xi_G + \sum_{h \in {\cal H}_G^-} z_{hk} & \geq 1 ~~ \qquad\forall k \neq k_G,~ \forall G \in {\cal G} \label{const:groupFalsePos} \\ 
M\xi_G - \sum_{h \in {\cal H}_G^+} z_{hk} & \geq 0 ~~\qquad \forall k = k_G,~ \forall G \in {\cal G} \label{const:groupFalseNeg} \\ 
\sum_{i \in {\cal G}} |G_i|\xi_i &\leq \alpha \label{const:error_bound_grouped}
\end{align}
where $k_G$ is the cluster of group $G$. Note that constraints \eqref{const:groupFalsePos} and \eqref{const:groupFalseNeg} are nearly identical to the non-grouped version except the sets of hyperplanes are now defined for hyper-rectangles. Constraint \eqref{const:error_bound_grouped} now weights the error of the solution by the size of the group.

To alter the pricing problem for the grouped setting we update the constraints that check whether or not a data point is correctly included in the half-space to check the entire hyper-rectangle. Specifically we update constraints \eqref{const:halfspace_error_1} and \eqref{const:halfspace_error_2} to the following:

\begin{align}
\hskip3cmw_+^T (x_G^{H}) - w_-^T (x_G^{L}) - b &\leq M\delta_i ~~&& \forall G \in {\cal G}_k  \label{const:group_halfspace_error_1}\\[.3cm]
w_+^T (x_G^{L}) - w_-^T (x_G^{H}) - b  &\geq \epsilon -M \delta_i  ~~&& \forall G \in {\cal G} \setminus {\cal G}_k \hskip3cm \label{const:group_ halfspace_error_2}
\end{align}

Note that groups in ${\cal G}_k$ are only correctly in the half-space if the entire box is included in the half-space, and similarly groups outside of ${\cal G}_k$ are only correctly outside the half-space if the entire box is outside the half-space.

% Simulation Results

\subsection{Empirical Evaluation} \label{sec:group_exp}

To evaluate the performance of our grouped data approach versus sub-sampling data points we ran a sequence of experiments on synthetic data. Data was generated using a Gaussian mixture model where cluster centers were sampled uniformly from $[-1,1]^m$, and $n$ data points were generated around the sampled center for each cluster with a covariance matrix of $\sigma I$ where $I$ is the $m \times m$ identity matrix. The parameter $\sigma$ controls the difficulty of the description problem as larger values of $\sigma$ lead to clusters with considerable overlap making perfect explanation unlikely. To construct the groups for our approach we use hierarchical clustering with a limit on the maximal linkage distance $\epsilon$, which is akin to setting a maximum diameter on the size of the groups. We tested a range of different $\epsilon$ values to get different number of groups. To provide a fair comparison between the two approaches we sub-sampled the same number of data points (uniformly at random) as the number of groups. The same set of candidate half-spaces, generated by considering all possible uni-variate splits, is also used for both approaches. For all of the following results we created 50 random instances using the above simulation procedure with $K=3$, $m=10$, and $n=10000$ and then ran both approaches and averaged the performance over the 50 instances, and 5 random sub-samples.

\begin{figure}[!bht]
    \centering
    \begin{subfigure}
      \centering
      \includegraphics[width=0.49\textwidth]{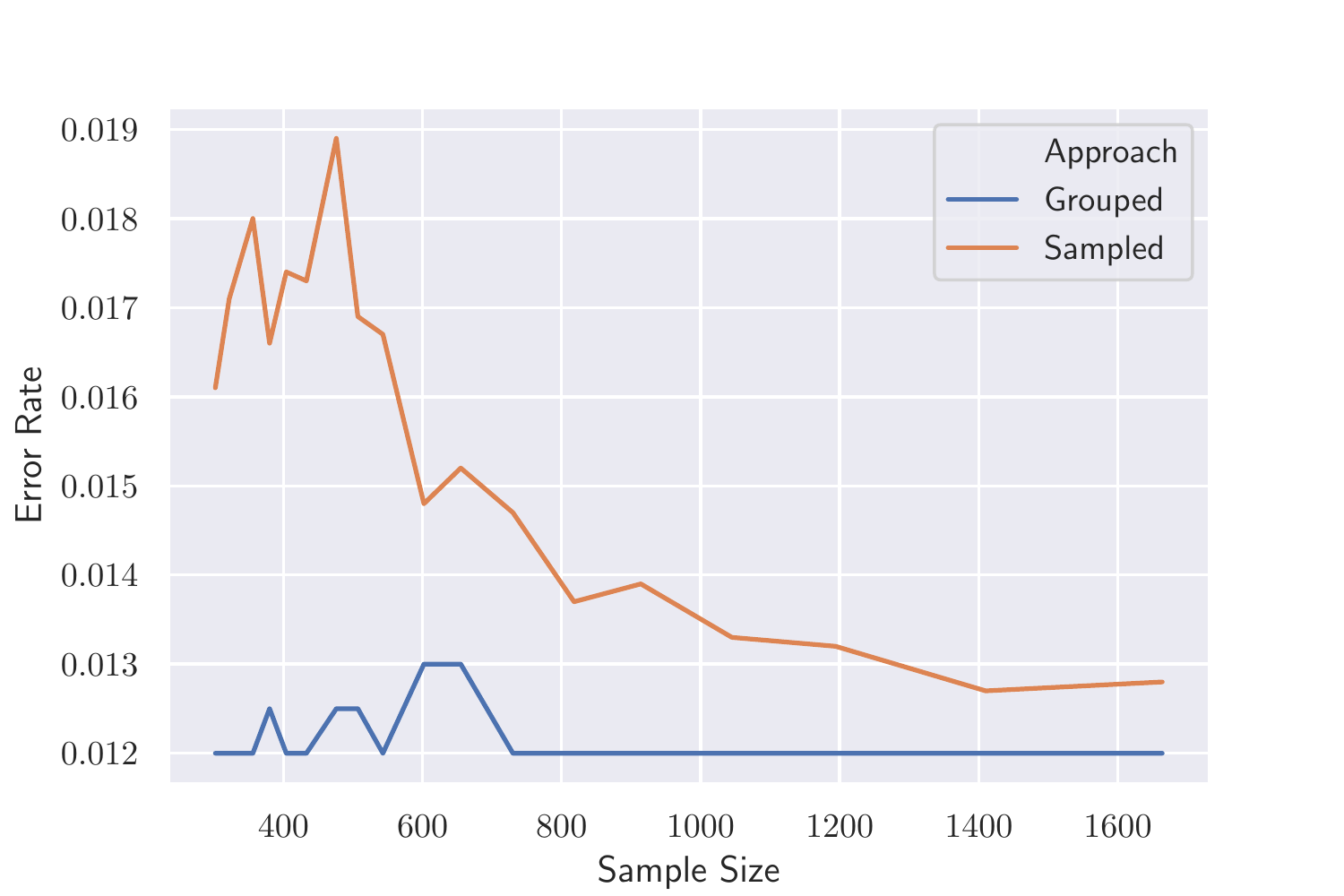}
    \end{subfigure}
    \begin{subfigure}
      \centering
      \includegraphics[width=0.49\textwidth]{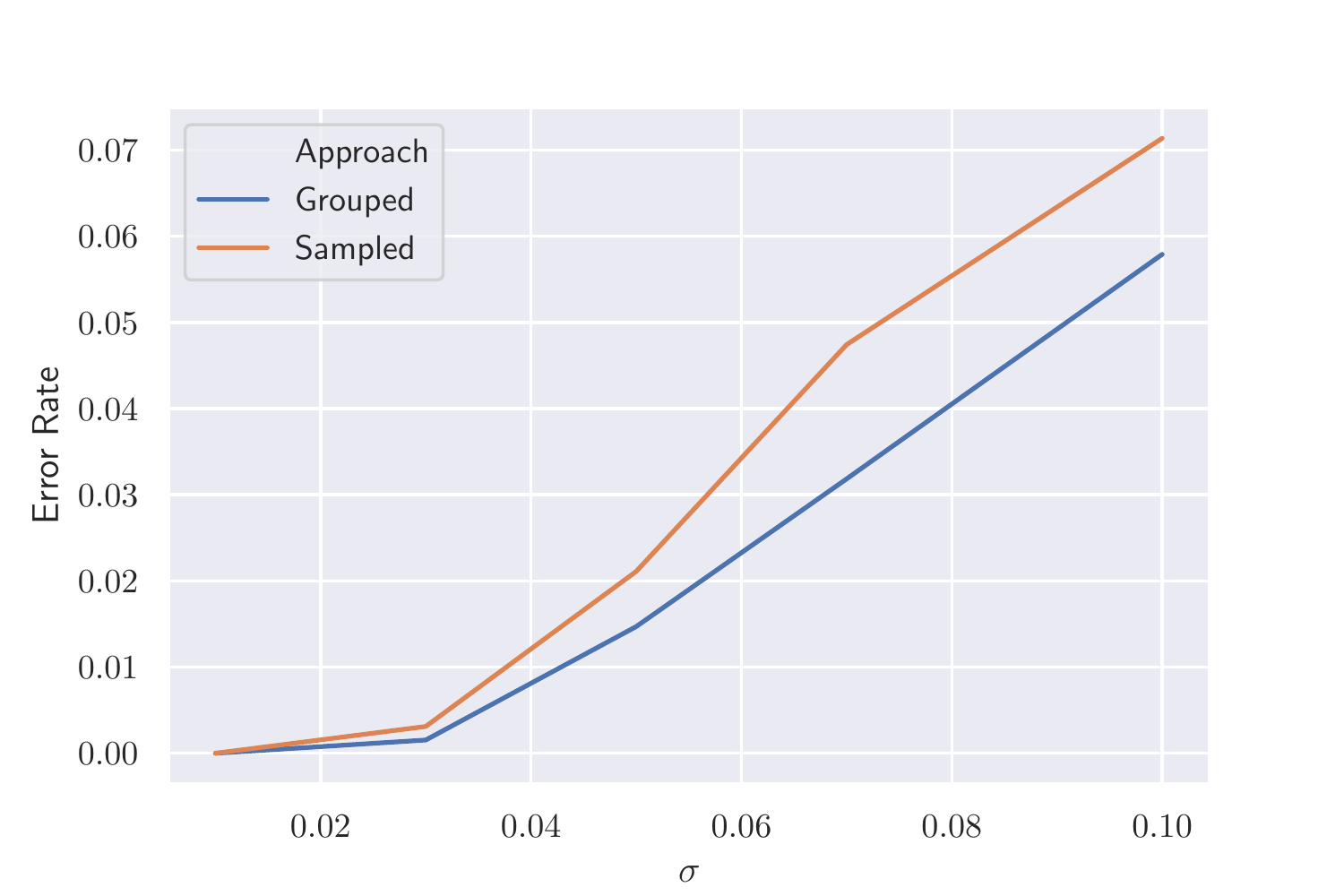}
    \end{subfigure}
  \caption {\label{synthetic_plots} Relative performance of grouping data versus sub-sampling. Error rate is what percentage of dataset is not properly explained by the explanation generated using each approach. Sample size is the number of groups or data points for the grouped and sub-sampled approach respectively. $\sigma$ is the standard deviation of the generated Gaussian clusters. The second plot used a sample size of 800.
  }
\end{figure}

Figure \ref{synthetic_plots} shows the results of the synthetic experiments. The results show that for an equivalent number of samples (i.e. groups for the grouped data and data points for the sub-sampled data) the grouping approach is able to find explanations with a lower error rate. This trend also holds as we increase the difficulty of the problem instances, with grouping achieving better performance at all choices for $\sigma$. Together this provides compelling empirical evidence that grouping is an effective tool for scaling our IP approach to larger data sets.

% Graphs

%Discussion of results

\section{Numerical Results} \label{sec:exp}
% Datasets + pre-processing
To evaluate our approach we ran experiments on a suite of clustering datasets from the UCI Machine Learning repository \citep{asuncion2007uci}. We pre-process all datasets by using a min-max scaler to normalize numeric feature values between 0 and 1, encode all categorical features using one-hot encoding, and for all supervised learning datasets remove the target variable. To create a reference cluster assignment we use $k$-means clustering using $k$-means++ initialization scheme with 100 random restarts. To select the number of clusters we tune $k$ between 2 and 10 and select the $k$ with the best silhouette score.     Note that for certain choices of $\alpha$ the MIP and MLP may be infeasible. As $\alpha$ is not given as a constraint for the application a priori in these datasets, we use a two-stage procedure to first find a feasible $\alpha$ then optimize for interpretability. In the first phase we replace the objective in the MIP \eqref{obj:mip} with $\alpha$ which we take as a continuous decision variable. The goal of the first stage is thus to optimize for the accuracy of the descriptions. We then take the optimal $\alpha^*$ from the first stage and multiply it by a tolerance factor (i.e. $(1 + \kappa)\alpha^*$ for a small $\kappa$) and use it in constraint \eqref{const:error_bound} in the second stage to optimize for interpretability of the descriptions. For the following experiments we used $\kappa = 0.05$.

% Benchmark algorithms
We benchmark our approach against three common algorithms for cluster description: Classification and Regression Trees (CART) \citep{breiman2017classification}, Iterative Mistake Minimization Trees (IMM) \citep{frost2020exkmc}, and Prototype Descriptions (PROTO) \citep{carrizosa2021clustering}. We do not compare against the Disjoint-Tag Minimization Model \citep{davidson2018cluster} as the approach requires data in a different form to the preceding algorithms. For all approaches we used the same $k$-means clustering as a reference cluster assignment to be explained. For CART we used the cluster assignments as labels for the classifier. For both CART and IMM we set the number of leaf nodes to be the number of clusters to provide a fair comparison to the polyhedral description approach. While IMM is an algorithm for generating new clusters not explaining the reference clustering, we interpreted the resulting tree as an explanation for the initial clustering. While in principle IMM should under-perform CART which explicitly optimizes for classification accuracy we found that IMM outperformed CART with respect to explanation accuracy on a number of datasets. We implemented the Prototype description IP model using Gurobi 9.1 \citep{gurobi} and Python, and placed a 300 second time limit on the solution time. To allow the prototype description model to scale to larger datasets we implemented the sub-sampling scheme outlined in the original paper and sub-sampled 125 candidate prototypes and 500 data points for each cluster.

% Our algo
We present results for both the low complexity (LC-PDP) and sparse (Sp-PDP) variants of our algorithm. We also consider two different settings for $\beta$ and $W$: PDP-1 which has $W=\beta=1$ and PDP-3 which has $W=10$, $\beta = 3$. For the following results, the pre-fix of the algorithm denotes the objective used and the suffix denotes the setting for $W$ and $\beta$. For instance, LC-PDP-1 refers to the low-complexity variant of our algorithm with $W=\beta=1$. To construct an initial set of candidate half-spaces, for each cluster we enumerate the $p$ maximum and minimum values for each each feature and construct half-spaces with uni-variate splits at each of the values. For instance, if the points in a cluster have values between 0 and 5 for feature $d$ for $p=1$ we add candidate half-spaces $\{x_d \geq 0\}$ and $\{x_d \leq 5\}$. For the following experiments we chose $p=10$. For all results we set a 300 second time limit on the overall column generation procedure and a 30 second time limit on solving an individual pricing problem. We add all solutions found during the execution of the pricing problem with negative reduced cost to the master problem. All models were implemented in python using Gurobi 9.1 and run on a computer with 16 GB of RAM and a 2.7 GHz processor. 

% Describe different result tables and why PROTO only shows up in one
Table \ref{tab:accuracy} shows the performance of each algorithm with respect to cluster description accuracy. Overall PDP is able to dominate the other benchmark algorithms, achieving the best accuracy on every benchmark dataset. Surprisingly, PDP-1 and PDP-3 perform almost identically, with PDP-3 only outperforming PDP-1 on the seeds dataset. Overall, PROTO is the least competitive approach, likely due to being the most restrictive function class relative to decision trees and polyhedra.

\begin{table}[!h]\small
\caption{Cluster description accuracy (\%). The percentage of data points in the original reference clustering that are correctly explained. Bolded numbers indicate best accuracy for each dataset.}
\centering
\label{tab:accuracy}
\begin{tabular}{lrrrrrrrrr}
\toprule
    Dataset &      n &  m & K &     IMM &    CART &   PROTO &  PDP-1 &   PDP-3 \\
\midrule
      adult &  32561 &   108 & 3 &   99.93 &   99.63 &   66.40 &     \textbf{99.95} &      \textbf{99.95} \\
       bank &   4521 &  51 & 7 &   \textbf{97.74} &   92.79 &    80.1 &     \textbf{97.74} &      \textbf{97.74} \\
    default &  30000 &  23 & 2 &  \textbf{100.00} &  \textbf{100.00} &   99.2 &    \textbf{100.00} &     \textbf{100.00} \\
      seeds &    210 &   7 & 2 &   98.57 &   98.57 &   98.10 &     99.05 &       \textbf{100.00} \\
        zoo &    101 &   17 & 4 &  \textbf{100.00} &  \textbf{100.00} &   95.05 &    \textbf{100.00} &     \textbf{100.00} \\
       iris &    150 &   4 & 2 &  \textbf{100.00} &  \textbf{100.00} &  \textbf{100.00} &    \textbf{100.00} &    \textbf{100.00} \\
 framingham &   3658 &   15 & 8 &  \textbf{100.00} &  \textbf{100.00} &    82.8 &    \textbf{100.00} &    \textbf{100.00} \\
       wine &    178 &  13 & 2 &   97.19 &   97.19 &   96.63 &     \textbf{98.88} &    \textbf{98.88} \\
     libras &    360 &  90 & 10 &   82.50 &   78.06 &   78.61 &     \textbf{98.06} &      \textbf{98.06} \\
       spam &   4601 &   57 & 2 &   \textbf{99.98} &   \textbf{99.98} &   94.07 &     \textbf{99.98} &      \textbf{99.98} \\
\bottomrule
\end{tabular}
\end{table}

Table \ref{tab:sparsity} shows the number of features used in the cluster descriptions. Note that PROTO does not appear in this table or the complexity table as the output for each cluster is simply a representative data point and a radius, and thus has no natural analog for sparsity or complexity. We report results for Sp-PDP as it directly optimizes this metric, whereas we report complexity for the LC-PDP. Sp-PDP performs competitively with IMM and CART getting the best sparsity in all but three datasets. Of the three datasets where it is outperformed by CART it is important to note that Sp-PDP achieves considerably better accuracy highlighting that the gains in explanation accuracy can come at a cost to the interpretability of the explanation. 

\begin{table}[!h]\small
\caption{Cluster description sparsity for explanation : number of features used in total across all half-spaces in description. Bolded numbers indicate best sparsity for each dataset.}
    \centering
\label{tab:sparsity}
\begin{tabular}{lrrrrrrrr}
\toprule
    dataset &      n &   m & K & IMM &  CART &  Sp-PDP-1 &  Sp-PDP-3 \\
\midrule
      adult &  32561 &   108 &   3 &  2 &     2 &       \textbf{1} &     \textbf{1} \\
       bank &   4521 &   51 & 7 &  6 &     6 &      \textbf{5} &     \textbf{5} \\
    default &  30000 &  23 & 2 &  \textbf{1} &     \textbf{1} &       \textbf{1} &       \textbf{1} \\
      seeds &    210 &   7 & 2 & \textbf{1}  &     \textbf{1} &       2 &     3 \\
        zoo &    101 &   17 & 4 &  \textbf{3}  &     \textbf{3} &      \textbf{3} &     \textbf{3} \\
       iris &    150 &   4 & 2 &  \textbf{1}  &     \textbf{1} &      \textbf{1} &       \textbf{1} \\
 framingham &   3658 &   15 & 8 &   \textbf{3}  &     \textbf{3} &      \textbf{3} &      \textbf{3}\\
       wine &    178 &   13 & 2 &   \textbf{1} &     \textbf{1} &      4 &      2 \\
     libras &    360 &  90 & 10 & \textbf{9} &     \textbf{9} &     18 &      18 \\
       spam &   4601 &   57 & 2  &    \textbf{1} &     \textbf{1} &       \textbf{1} &       \textbf{1} \\
\bottomrule
\end{tabular}
\end{table}

Finally, Table \ref{tab:complex} shows the complexity of the resulting cluster descriptions. For CART and IMM we compute the complexity by considering each internal branching node as a half-space and report the total complexity of half-spaces needed to explain each cluster, counting a half-space multiple times if it is used to describe multiple clusters to provide a fair comparison to polyhedra. LC-PDP also performs competitively with the decision tree based approaches only being outperformed on datasets where it achieves higher cluster accuracy.

\begin{table}[!t]\small
    \centering
    \caption{Complexity of Cluster Description. Bolded numbers indicate best complexity for each dataset.}
    \label{tab:complex}
\begin{tabular}{lrrrrrrrr}
\toprule
    Dataset &      n &  m &  K &  IMM &  CART &  LC-PDP-1 &  LC-PDP-3 \\
\midrule
      adult &  32561 &   108 & 3 &    \textbf{10} &     \textbf{10} &      \textbf{10} &     \textbf{10} \\
       bank &   4521 &   51 & 7 &    44 &    42 &      \textbf{40} &      \textbf{40} \\
    default &  30000 &   23 & 2 &    \textbf{4} &     \textbf{4} &       \textbf{4} &       \textbf{4} \\
      seeds &    210 &   7 & 2 &    \textbf{4} &     \textbf{4} &       \textbf{4} &       \textbf{4}\\
        zoo &    101 &   17 & 4 &    18 &     18 &       \textbf{14} &       \textbf{14} \\
       iris &    150 &   4 & 2 &    \textbf{4} &     \textbf{4} &       \textbf{4} &       \textbf{4} \\
 framingham &   3658 &   15 & 8 &   48 &    48 &      48 &       \textbf{44} \\
       wine &    178 &   13 & 2 &    \textbf{4} &     \textbf{4} &      10 &   6 \\
     libras &    360 &  90 & 10 &   98 &    82 &      84 &      \textbf{80} \\
       spam &   4601 &  57 & 2 &    \textbf{4} &     \textbf{4} &       \textbf{4} &    \textbf{4} \\
\bottomrule
\end{tabular}
\end{table}

\begin{figure}[!htb]

    \centering
    \begin{subfigure}
    \centering
      \includegraphics[width=0.3\textwidth]{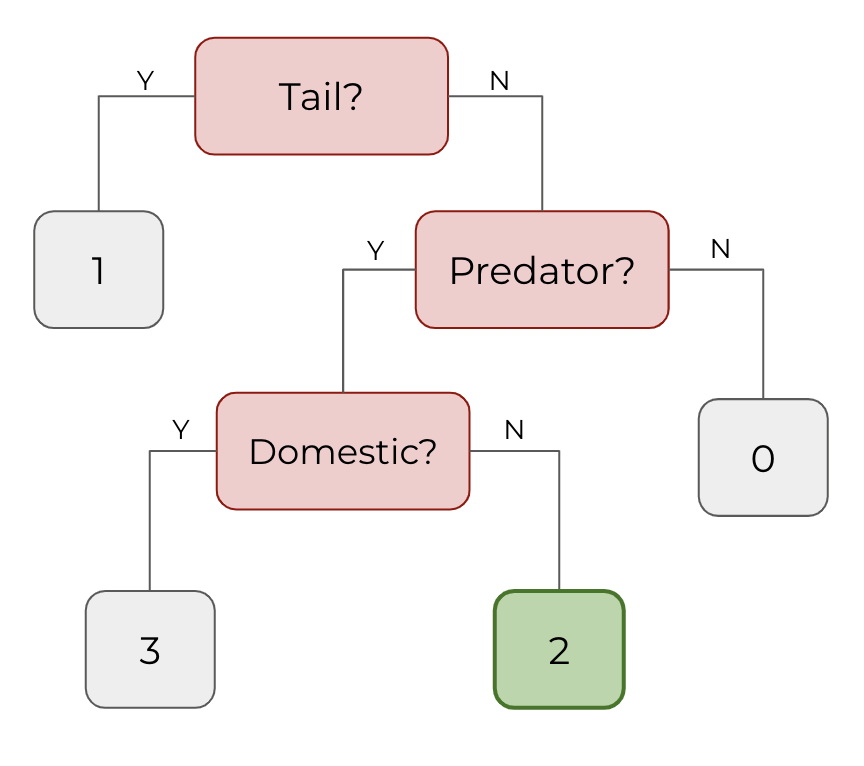}
    \end{subfigure}
    \begin{subfigure}
    \centering
      \includegraphics[width=0.3\textwidth]{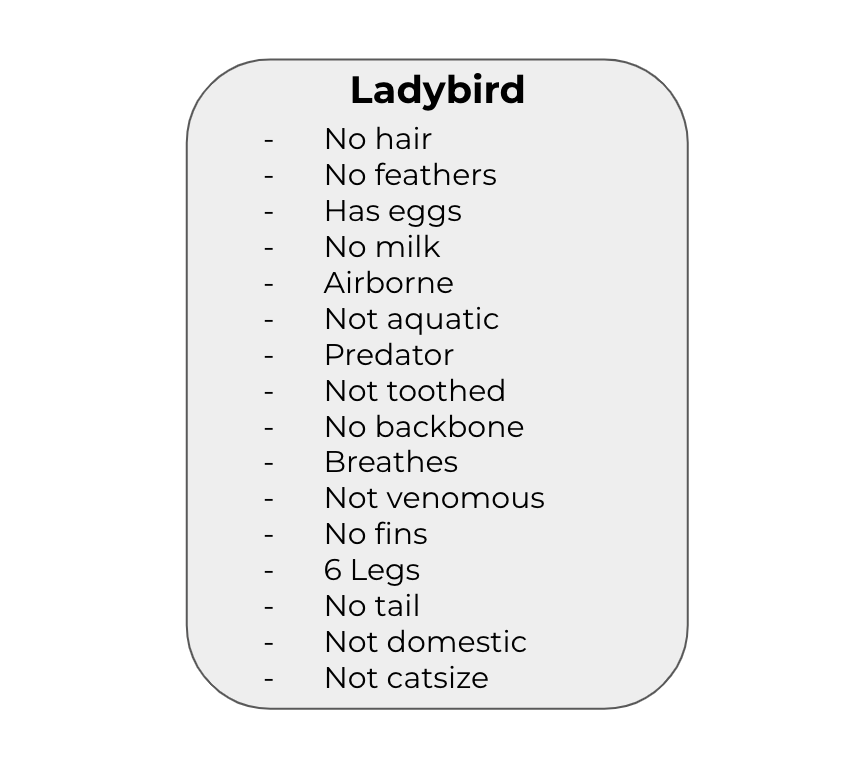}
    \end{subfigure}
    \begin{subfigure}
    \centering
      \includegraphics[width=0.3\textwidth]{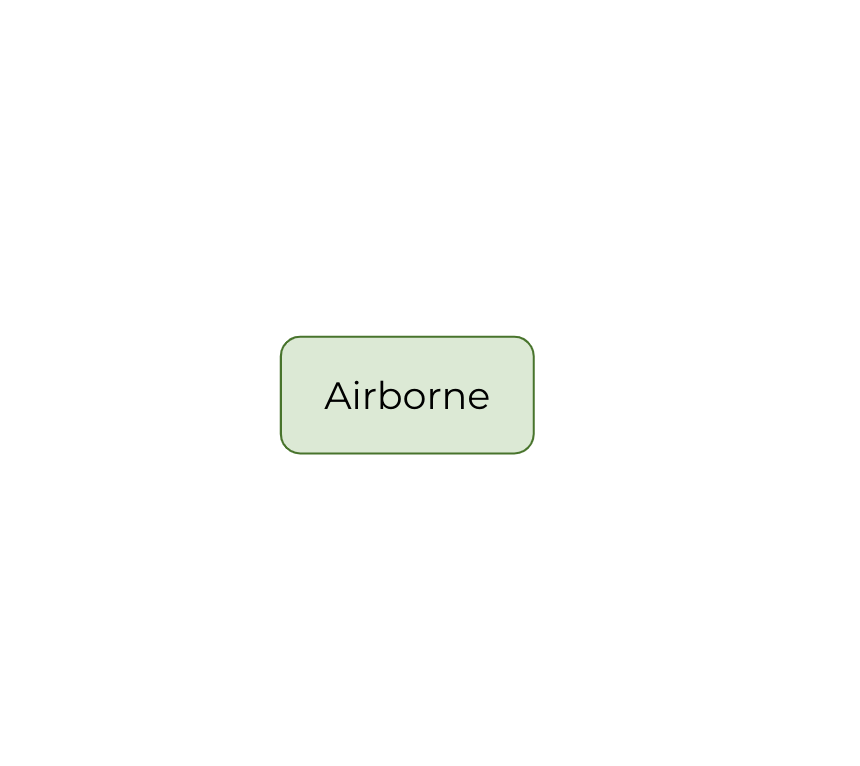}
    \end{subfigure}
  \caption {\label{fig:interpretability_ex} Sample cluster descriptions for the same cluster on the zoo dataset. (Left) A decision tree. (Middle) A prototype. (Right) A polyhedral description.} 
\end{figure}

Figure \ref{fig:interpretability_ex} shows three sample cluster descriptions for the zoo dataset to compare each model class's interpretability. For this example we use the best reference k-means clustering which resulted in four clusters, and describe the second cluster (which is composed primarily of birds). The prototype explanation for the cluster is a ladybird. While having a representative animal is easy to understand, without the added context that the cluster is primarily birds it is not obvious what are the defining characteristics of the cluster. For instance, ladybirds are also predators and have eggs, which could also define clusters. The decision tree description requires that the cluster has no tail, is a predator, and is not domestic. Compared to both the decision tree and prototype explanation, the polyhedral description, simply that the cluster is all airborne, provides a parsimonious summary of the cluster that gives intuition about its defining characteristic. This further underscores that a full partition of the feature space for a description, as necessary for a decision tree, may lead to more complicated descriptions.

\begin{comment}
\begin{figure}[!htb]

    \centering
    \begin{subfigure}
    \centering
      \includegraphics[width=0.8\textwidth]{figures/interpretability_ex.png}
    \end{subfigure}
  \caption {\label{fig:interpretability_ex} (Left) A sample set of clusters to be explained where the convex hulls do not intersect and perfect explanation is possible. (Middle) Polyhedral description using convex hull of clusters. (Right) Lower complexity polyhedral description.} 
\end{figure}

\end{comment}

\section{Conclusion}
In this paper we introduced a novel approach for cluster description that works by describing clusters with a polyhedron. As opposed to existing approaches, our algorithm is able to explicitly optimize for the complexity or sparsity of the resulting explanations. We formulated the problem as an integer program and present both a column generation procedure to deal with an exponential number of candidate half-spaces and a grouping scheme to help the approach scale to large datasets. Compared to state of the art cluster description algorithms our approach is able to achieve competitive performance in terms of explanation accuracy and interpretablity when measured by sparsity and complexity.

Our method currently only leverages a single polyhedron per cluster description but a promising direction for future work is extending the framework to allow for multiple polyhedra. A natural extension could involve allowing each cluster to leverage multiple polyhedra with a budget on the total number of polyhedra used across cluster descriptions, akin to a constraint on the total number of leaf nodes in a decision tree.

Currently our method is agnostic to the final polyhedra selected as long as the interpretability and accuracy performance is the same. However, in many applications it could be desirable to either have descriptions that are as compact as possible (i.e. the polyhedron closely captures the given cluster) or descriptions that are broader (i.e. for potentially new data). As a post-processing step, the right hand sides of the half-spaces $b$ that compose each polyhedron can be adjusted as needed for the application. For instance, for applications where a compact description is desirable, each half-space could be shrunk until it hits a data point in the cluster. We leave as future work an approach to make the given polyhedra broader for new data.

\bibliography{main}
\bibliographystyle{plain}

\appendix
\section{Appendix}

\subsection{Illustrative Example on Non-Intersection of Polyhedral Descriptions}
Consider a simple example where we have two clusters representing dogs and cats and two binary features - one indicating whether the animal barks and the other if it meows. A simple polyhedral description for these clusters is $\text{BARKS}=\text{TRUE}$ for the dog cluster and $\text{MEOWS}=\text{TRUE}$ for the cat cluster. However, the two polyhedra intersect in the improbable region where an animal both barks and meows. For this simple example, a solution could be to use $\text{BARKS}=\text{FALSE}$ for the cat cluster. However, if we increase the number of animals, each with their own new binary feature for the noise they make (i.e. a frog cluster with a binary feature for ribbets), then our polyhedral descriptions can either intersect, with one simple half-space per cluster, or the description needs to add additional conditions (i.e. $\text{BARKS}=\text{FALSE}$ and $\text{RIBBETS}=\text{FALSE}$ for the cat cluster) which make the resulting description harder to interpret solely for scenarios that are unlikely to occur in real world data. 

\subsection{Proof of Theorem \ref{thm:np} }
\begin{proof}
We start by noting that membership in NP is straightforward (i.e. given a solution it is easy to see whether or not the given polyhedra correctly explain the given clusters). We'll prove NP-Hardness by a reduction from 3-SAT. Consider a 3-SAT problem with $n$ variables $v_1, v_2, \dots, v_n$ and $m$ clauses $K_1, K_2, \dots K_m$. Each clause $K_i$ consists of three conditions $(v_{i1} \lor v_{i2} \lor v_{i3})$  where $v_{ij}$ corresponds to either one of the original variables or its complement. We'll now construct a LC-PDP instance with $2n$ candidate half-spaces in $2n$ dimensional feature space with $m + n + 1$ data points. We focus specifically on the simplest form of the problem - explaining only one cluster. Clearly if explaining one cluster is NP-Hard, explaining multiple cluster will also be NP-Hard. Let ${\cal C}_0$ be the cluster to be explained.

For each original variable $v_i$ we add two new dimensions $d_{v_i}$ and $d_{\bar{v}_i}$. We also add two candidate half-spaces $h_{v_i} = \{x: x_{d_{v_i}} \leq 0.5\}$ and $h_{\bar{v}_i} = \{x: x_{d_{\bar{v}_i}} \leq 0.5\}$. We generate one data point $x^0$ in ${\cal C}_0$ that has a value of 0 for every feature. For each variable $v_i$ in the original 3-SAT problem we add one new data point $x^{v_i}$ outside the cluster to be explained that has 1s for features $d_{v_i}$ and $d_{\bar{v}_i}$, and 0s otherwise. This adds a total of $n$ new data points. We also add one data point $x^{K_i}$for every clause $K_i$ in the original 3-SAT problem, which has 1 for the features corresponding to the original conditions in the clause $d_{v_{i1}}, d_{v_{i2}}, d_{v_{i2}}$ and 0s otherwise. For instance if the original clauses was $v_1$ or ${\bar v}_2$ or $v_3$, then the associated data point in the PDP would have 1s for features $d_{v_1}, d_{\bar{v}_2}, d_{v_3}$ and 0s otherwise. This adds a total of $m$ new data points bringing the total number of data points to $n + m + 1$. Finally we add a complexity constraint to the instance of $2n$. Note that because each half-space uses one feature, this is equivalent to adding a constraint that at most $n$ half-spaces can be used.

The above instance can clearly be set-up in polynomial time. It now suffices to show that solving the associated PDP yields a valid solution to the 3-SAT problem. 

We start be claiming that the solution to the aforementioned LC-PDP yields solutions where exactly one of $h_{v_i}$ and $h_{\bar{v}_i}$ are used. Consider if this were not true. Then the solution to the LC-PDP must have a solution where either both $h_{v_i}$ and $h_{\bar{v}_i}$ or neither are. However, at least one of $h_{v_i}$ and $h_{\bar{v}_i}$ must be used, otherwise $x^{v_i}$ would not be classified correctly. We also know that a feasible solution cannot use multiple half-spaces corresponding to one variable, given that each variable has at least one half-space used, because it would contradict the complexity constraint that at most $n$ half-spaces used. Thus the claim must be true.

Given the above claim, we can now interpret the half-spaces selected as the variable settings in the original 3-SAT problem (i.e. $v_i = T$ if $h_{v_i}$ is selected and $v_i = F$ if $h_{\bar{v}_i}$ is selected). We now claim that a solution to the LC-PDP corresponds to a solution of the 3-SAT instance. Note that by the feasibility of the LC-PDP solution we have that for each data point outside ${\cal C}_0$ there exists at least one half-space selected that excludes it. By construction we know for every clause in the original 3-SAT problem there is an associated data point $x^{K_i}$ outside the cluster to be explained that is only excluded by the half-spaces corresponding to the conditions in the clause $h_{v_{1i}}, h_{v_{2i}}, h_{v_{13}}$. Thus at least one of the half-spaces corresponding to the conditions must be used, and by extension every clause must be satisfied. An identical proof also works if we replace the complexity constraint with a sparsity constraint (as each half-space uses a new dimension) thus also completing the claim for Sp-PDP.
\end{proof}

\subsection{Proof of Theorem \ref{thm:group_ub}}
\begin{proof}
We start by noting some properties of $COST_{G}(\textbf{P})$ and $COST(\textbf{P})$. First, for a fixed solution $\textbf{P}$ $COST_{G}(\textbf{P}) \geq COST(\textbf{P})$ - which follows directly from the fact that the grouped cost over-estimates error (i.e. counts all members of group as mis-classified if any individual data point in the group is mis-classified). By the definition of $\textbf{P}^*_{G}$, and $\textbf{P}^*$ we also have that $COST_G(\textbf{P}^*_{G}) \leq COST_G(\textbf{P}^*)$ and $COST(\textbf{P}^*) \leq COST(\textbf{P}^*_G)$ respectively. Rearranging the three inequalities we get:

$$
COST_G(\textbf{P}^*) \geq COST(\textbf{P}^*_G) \geq COST(\textbf{P}^*)
$$

This implies that if we can get a bound on the difference between the grouped cost and full cost of $\textbf{P}^*$ we can get a bound on the sub-optimality of $\textbf{P}^*_G$ for the full problem.

Take $\textbf{P}^*$ and consider the grouped cost relative to the original cost. Looking at each group $G$ individually there are three possible cases: All the data points in a group are correctly classified, all data points in the group are misclassified, and the group has both data points that are both classified correctly and incorrectly. In the former two cases, the grouped cost is identical to the original cost, so it suffices to consider the last case. Note that the additional increase in cost for that group is equal to the number of correctly classified data points in the group. In the worst case, there are at most $|G| - 1$ such points. Thus, the cost in the grouped setting is at most $|G|$ times the original cost for data points in that group. Overall, in the worst case this is the only case in the dataset and every group it affects is the largest possible size $|G_{max}|$ completing the claim that the overall grouped cost is $|G_{max}|$ times the original cost completing the result. Note that no aspect of the proof uses how the groups were constructed, so the result holds for any grouping scheme.
\end{proof}

\subsection{Proof of Theorem \ref{thm:group_lb}}

\begin{proof}
Consider a simple example with two clusters and a single feature $x$. For the first cluster $C_0$ there are three data points at the origin ($x = 0$) and m data points placed individually at increments of $-d_2$ (i.e. one data point at $x=-d_2$, one data point at $x=-2d_2$ and so on). For the second cluster $C_1$ there is one data point at the origin, 2 data points at $x=d_1$, and $m$ data points placed at increments of $d_2$ after $d_1$ (i.e. one data point at $x=d_1+d_2$, one data point at $x=2d_2+d_1$ and so on). We set $d_1 < d_2$. Figure \ref{fig:thm32visual} shows a visualization of the setting.

\begin{figure}[h]
    \centering
    \begin{subfigure}
    \centering
      \includegraphics[width=12cm]{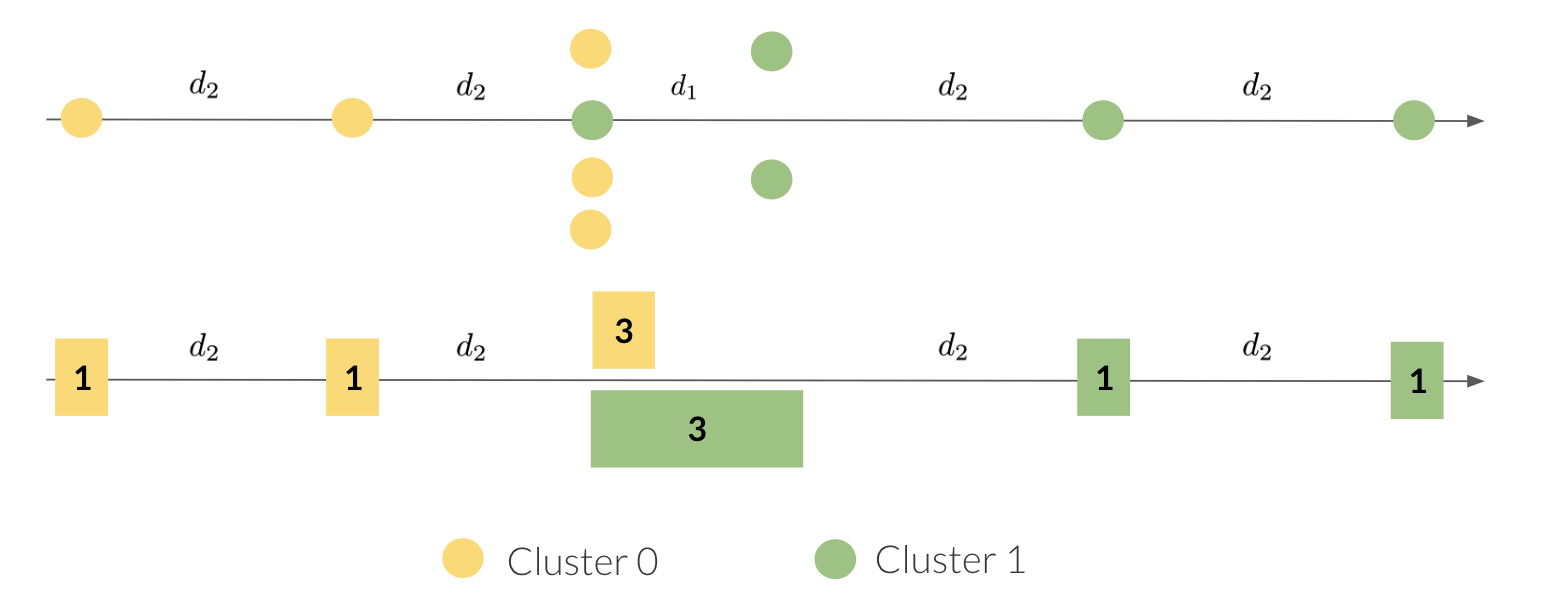}
    \end{subfigure}
  \caption {\label{fig:thm32visual} (Top) Visualization of points to be explained in instance for  Theorem \ref{thm:group_lb}. (Bottom) Optimal grouping with respect to silhouette coefficient for $|G_k| = |C_k| - 1$. } 
\end{figure}

Consider the following groupings which we claim are optimal with respect to the silhouette coefficient. For $C_0$ all three data points at the origin form one group and every other data point is in its own group. Evidently this is the optimal grouping for $|G_0| = |C_0| - 2$ as every group has an intra-cluster distance of 0 and an inter-cluster distance of $d_2$ giving a silhouette score for the grouping of 1. For $C_1$ we group the one data point at the origin and the 2 data points at $x=d_1$ together, and every other data point is in its own group. Suppose this was not optimal with respect to the silhouette coefficient for $|G_1| = |C_1| - 2$. Clearly an optimal grouping will have the two points at $x=d_1$ together as they have an intra-cluster distance of 0. Thus the only scenarios are that the point at $x=d_1+d_2$ is included in that group or two of the $m$ points spaced at increments of $d_2$ are grouped together. Simple arithmetic shows that both scenarios result in a silhouette coefficient larger than the given grouping, proving its optimality.

An optimal solution to the original problem is to use a single half-space $\{x \in \mathbbm{R}: x \leq 0\}$ for $C_0$ and $\{x \in \mathbbm{R}: x \geq d_1\}$ for $C_1$ respectively, which incurs a cost of 1. Note that under the optimal grouping scheme outlined above one group with 3 points from $C_0$ overlaps with one group with 3 points from $C_1$. Thus an optimal solution to the grouped problem is to use a single half-space $\{x \in \mathbbm{R}: x \leq d_1\}$ for $C_0$ and $\{x \in \mathbbm{R}: x \geq d_1+\epsilon\}$, where $\epsilon < d_2$, for $C_1$ respectively as no solution will incur a grouped cost less than 3. This optimal solution to the grouped problem incurs a true cost of 3 (as the three points in 3 point group in $C_1$ are mis-classified), completing our claim.

%Consider a simple example with two clusters and a single feature $x$. The first cluster has $n$ data points with feature value 0, and a single data point, denoted the bad data point, with feature value 1. The second cluster has $n$ data points with feature value 1. There are two candidate half-spaces $h_1 = \{x: x \leq 0.5\}$ and $h_2 = \{x: x \geq 0.5\}$. The optimal solution to the full version of the problem is to have cluster 1 explained by $h_1$ and cluster 2 explained by $h_2$, which has a cost of 1 for the data point from cluster one that cannot be separated from cluster two. However note that under a grouping scheme any group containing the bad data point in cluster 1 will be misclassified. This gives a grouped cost equal to the size of the group containing the bad data point. Taking the worst case where that group is the largest group we get the desired result.
\end{proof}

\end{document}